\documentclass{svproc}

\usepackage{graphicx}
\usepackage{amssymb}
\usepackage{color,soul}
\usepackage{lineno}
\usepackage[ruled,vlined,linesnumbered,titlenumbered]{algorithm2e}
\usepackage{amsmath}
\usepackage{esvect}
\usepackage{times}
\usepackage{multirow}
\usepackage{enumitem}
\usepackage{amsbsy}
\usepackage{amssymb}
\usepackage{makecell}
\usepackage[dvipsnames]{xcolor}
\newcolumntype{L}[1]{>{\raggedright\let\newline\\\arraybackslash\hspace{0pt}}m{#1}}
\newcolumntype{C}[1]{>{\centering\let\newline\\\arraybackslash\hspace{0pt}}m{#1}}
\newcolumntype{R}[1]{>{\raggedleft\let\newline\\\arraybackslash\hspace{0pt}}m{#1}}
\newcommand\crule[3][black]{\textcolor{#1}{\rule{#2}{#3}}}
{ \newcommand{\mynote}[2]{
		\fbox{\bfseries\sffamily\scriptsize#1}
		{\small$\blacktriangleright$\textsf{\textcolor{red}{{\em #2}\bf }}$\blacktriangleleft$}}}
{ \newcommand{\mynote}[2]{}}
\usepackage{tabularx,ragged2e,booktabs,caption}
\newcolumntype{C}[1]{>{\Centering}m{#1}}

\begin{document}
\mainmatter              % start of a contribution
\title{A Quantum Computing-based System for Portfolio Optimization using Future Asset Values and Automatic Reduction of the Investment Universe}
\titlerunning{A Quantum Computing-based System for Portfolio Optimization}  % abbreviated title (for running head)

\author{Eneko Osaba\inst{1,3} 
	\and Guillaume Gelabert\inst{2} 
	\and Esther Villar-Rodriguez\inst{1}
	\and \\Antón Asla\inst{2}
	\and Izaskun Oregi\inst{1}}
\authorrunning{Osaba et al.} % abbreviated author list (for running head)
\tocauthor{Eneko Osaba, Guillaume Gelabert, Esther Villar-Rodriguez, \\ Antón Asla and Izaskun Oregi}
\institute{TECNALIA, Basque Research and Technology Alliance (BRTA), 48160 Derio, Spain.\\
\email{eneko.osaba@tecnalia.com},
\and
Serikat - Consultoría y Servicios Tecnológicos, 48009 Bilbao, Spain.
\and
Corresponding Author}

\maketitle              % typeset the title of the contribution

\pagenumbering{gobble}

\begin{abstract}
One of the problems in quantitative finance that has received the most attention is the portfolio optimization problem. Regarding its solving, this problem has been approached using different techniques, with those related to quantum computing being especially prolific in recent years. In this study, we present a system called \textit{Quantum Computing-based System for Portfolio Optimization with Future Asset Values and Automatic Universe Reduction} (\texttt{Q4FuturePOP}), which deals with the Portfolio Optimization Problem considering the following innovations: \textit{i}) the developed tool is modeled for working with future prediction of assets, instead of historical values; and \textit{ii}) \texttt{Q4FuturePOP} includes an \textit{automatic universe reduction} module, which is conceived to intelligently reduce the complexity of the problem. We also introduce a brief discussion about the preliminary performance of the different modules that compose the prototypical version of \texttt{Q4FuturePOP}.
\end{abstract}
\section{Introduction}
\label{sec:intro}
The present work aims to describe a quantum computing (\textit{QC}, \cite{gyongyosi2019survey}) based system for solving the portfolio optimization problem (\textit{POP}, \cite{thakkar2021comprehensive}). Briefly explained, the \textit{POP} intends to find the optimum asset allocation with the objective of \textit{i}) maximizing the expected return and \textit{ii}) minimizing the financial risk. More specifically, and following the Markowitz \textit{POP} formulation, the financial risk is calculated based on the diversification of the built portfolio \cite{becker2015markowitz}. Following this philosophy, the system tends to distribute the whole budget into different and uncorrelated assets rather than investing large amounts of money into the highest expected, albeit correlated, returns.

Formally described, the problem to be solved counts with \textit{i}) a group of $N$ assets $\mathcal{A}=\{a_0,\dots, a_i, \dots, a_{N-1}\}$; \textit{ii}) a dataset $AD = \{ad_0,\dots, ad_i, \dots, ad_{N-1}\}$, in which $ad_{i}$ is a list of historical values $ad_{i,k}$ of an asset $a_i$, representing $k$ a specific day within the complete period of $K$ days, and \textit{iii}) a total $bd$ budget.

Thus, the objective is to find the most promising assets in which to invest this budget, considering that i) all $bd$ must be invested and ii) the proportion of $bd$ that can be allocated to each asset is conditioned by the variable $p$. More concretely, and considering $w_i$ as the proportion of $bd$ invested on the asset $a_i$, this $w_i$ can be represented as the summation of any proportions $p_i$ in $P = \{p_0=bd,p_1=bd/2, ... p_{p-1}=bd/2^{p-1}, 0\}$, Furthermore, it should be deemed that $w_i<=bd$. With this notation in mind, the goal is to find the $W=\{w_0, w_1,...,w_{N-1}\}$ that maximizes the total expected return while minimizing the financial risk.

The system described in this paper, coined \textit{Quantum Computing-based System for Portfolio Optimization with Future Asset values and automatic universe reduction} (\texttt{Q4FuturePOP}), is a QC-based scheme for dealing with the $POP$ considering the following innovations:

\begin{itemize}
	\item \textit{Future projected values}: most of the QC based techniques proposed in the literature solve the $POP$ using as input historical dataset values for $\mathcal{A}$ \cite{herman2022survey}. In other words, developed solvers select the appropriate $W$ considering the past values of the group of available assets. On the contrary, for \texttt{Q4FuturePOP}, the input dataset is composed of future predictions to try to account for realistic environments in the $POP$ formulation. Using this input, \texttt{Q4FuturePOP} builds the complete dataset used for formulating the $POP$ problem considering future projected values of $\mathcal{A}$. Thus, the $W$ chosen by the system is based on future predictions instead of historical values. 
	
	\item \textit{Automatic universe reduction}: in order to calculate $W$ in a more efficient way, a search space reduction mechanism has been implemented in \texttt{Q4FuturePOP}. This mechanism works as follows: first, \texttt{Q4FuturePOP} takes as input the complete universe of $\mathcal{A}$. With this set of assets, the system conducts a number $E$ of preliminary executions which are used by \texttt{Q4FuturePOP} for detecting a subgroup of promising assets. After these preliminary executions, \texttt{Q4FuturePOP} builds an alternative $\mathcal{A}'$, which is a subgroup of $\mathcal{A}$ ($\mathcal{A}'\subseteq \mathcal{A}$). Using this newly generated $\mathcal{A}'$, the problem is finally executed, and the obtained outcomes are returned to the user. Thanks to this procedure, the complexity of the problem to solve is automatically decreased, allowing the system to reach a higher level of accuracy. 
\end{itemize}

The rest of the paper is structured as follows. Section \ref{sec:related} presents a brief overview of the background related to QC and POP. In Section \ref{sec:inout}, the inputs and outputs of \texttt{Q4FuturePOP} are described for the sake of understandability. After that, in Section \ref{sec:Q4FuturePOP}, the whole system is described. Then, in Section \ref{sec:discussion}, we discuss the preliminary performance of \texttt{Q4FuturePOP}. Section \ref{sec:conc} finishes this work by outlining some of the planned future work.

\section{Related Work} \label{sec:related}

The first POP-focused paper including real quantum experiments was published in 2015, in which the authors solved the problem using a prototype of the D-Wave's quantum annealer \cite{rosenberg2015solving}. With only 512 qubits at their disposal, the authors worked with a pool of 15 assets in which to invest. A second study focused on this topic was presented in 2017, exploring a specific investment case related to the Abu Dhabi Securities Exchange \cite{elsokkary2017financial}. In the following years, some interesting theoretical papers appeared, exploring different formulations and their possible resolutions using quantum approaches. Examples of this scientific trend can be found in \cite{rebentrost2018quantum} and \cite{kerenidis2019quantum}. These papers, published in 2018 and 2019, respectively, theorize the implementation of two algorithms without actually testing them on real quantum devices. 

Also in 2019, some advanced approaches were presented. In \cite{venturelli2019reverse}, for example, the authors employ a hybrid algorithm in which the quantum module is executed by the D-Wave 2000Q computer in order to improve the solutions found by a classical greedy algorithm. In the same year, the first paper focused on gate-based quantum computers was published, in which the authors present a \textit{Quantum Approximate Optimization Algorithm} \cite{hodson2019portfolio}. 

Particularly interesting are the papers \cite{cohen2020portfolio1} and \cite{cohen2020portfolio2}, published by \textit{Chicago Quantum} in 2020. In the first of these papers, the authors demonstrated how a hybrid solver can promisingly solve portfolios of up to 33 assets. In \cite{cohen2020portfolio2} a similar approach is proposed, in which the number of assets considered rises to 60 using another hybrid technique. Finally, in \cite{cohen2020picking}, the same authors managed to solve problems with a size of 134 assets, making use of the D-Wave Advantage System, composed of 5436 qubits.

Since 2021, the study of the POP through the quantum paradigm has experienced a significant increase. In \cite{mugel2021hybrid}, for example, a variant of the problem known as \textit{minimum holding time} is solved by the D-Wave quantum annealer, considering a universe of 50 assets in which to invest. In \cite{palmer2021quantum}, the authors present a problem in which different investment bands are deemed, allowing the fixing of a maximum permissible risk. Furthermore, a quantum-gate-based method was presented in \cite{yalovetzky2021nisq}, where a hybrid algorithm called \textit{NISQ-HHL} is proposed.

The study presented in \cite{grant2021benchmarking} is especially interesting for this paper. That work consists of a detailed analysis of the parameterization of the D-Wave's annealer. To do so, the authors use POP as a benchmarking problem, presenting a formulation of the problem that allows an investment granularity adapted to the user's needs. This study has proved to be interesting from a mathematical point of view since the formulation employed in this study is the one embraced in our work. Finally, it is worth highlighting the study presented in \cite{mugel2022dynamic}, where different quantum solvers are proposed based on both the quantum gate paradigm and the annealer. The authors of that work carried out different tests in a dynamic environment, solving problems with up to 52 assets.

As can be seen, the research conducted in recent years has been prolific. This section has attempted to briefly outline this vibrant activity. Being aware that the full state of the art is much broader, we refer interested readers to works as \cite{herman2023quantum}.

\section{Inputs and Outputs of \texttt{Q4FuturePOP}}\label{sec:inout}

Now, let's define some notation and terms in order to properly understand how \texttt{Q4FuturePOP} operates. From now on, we use the superscripts $^{h}$ for the historical data and $^{f}$ for the predicted future data. Furthermore, the daily return of an asset $a_i$ at day $k$ is $er^h_{k, i}=(ad^h_{k, i}-ad^h_{k-1, i})/ad^h_{k-1, i}$, resulting in $er^h\in\mathbb{R}^{K\times{N}}$ representing the daily returns of $\mathcal{A}$ in $AD^h$.

Thus, \texttt{Q4FuturePOP} receives as inputs \textit{i}) $AD^{h}\in\mathbb{R}^{K\times{N}}$, which contains all the historical values of $\mathcal{A}$ assets during $K$ days, \textit{ii}) a list $V=\{v_0, v_1,...,v_{N-1}\}$, in which $v_i$ is the initial value of $a_i$ at the time \texttt{Q4FuturePOP} is executed (in most cases, $v_i=ad^h_{K-1, i}$); and \textit{iii}) $Er^f=\{Er^f_0, Er^f_1,...,Er^f_{N-1}\}\in\mathbb{R}^{N}$, which is the list of predicted expected returns. %, where $Er^f_{i} = (ad^f_{L-1, i}-v_{i})/v_{i}$ is the predicted expected return for asset $a_{i}$ at the end of the upcoming $L-1$ days. 
It should be highlighted that these $Er^{f}$ values are given by an expert from the Spanish company Welzia Management\footnote{https://wz.welzia.com/}. 

Once the input is received, the module named \textit{Predicted Dataset Generation} (\texttt{PDG}) builds the complete dataset $AD^{f}\in\mathbb{R}^{K\times{N}}$, which is composed of all the projected future daily values of the whole $\mathcal{A}$ in the period that goes from the moment in which the system is executed and the following $K$ days. All values in the generated $AD^f$ must meet two requirements: \textit{i}) $Cov(er^{f}) = Cov(er^{h})$, meaning that the covariance of the input daily returns is the same as that of the daily returns generated by \texttt{PDG}; and \textit{ii}) the expected return of the prices build must be the same as $Er^{f}$. How $AD^f$ is generated is described in Section \ref{sec:generation}.

As output, \texttt{Q4FuturePOP} returns \textit{i}) the above-mentioned $W$, containing the list of investment weights $w_i$ given to each $a_i$; \textit{ii}) the expected return of the chosen portfolio $er_{portfolio}$; \textit{iii}) the risk associated with this portfolio $\sigma_{portfolio}$; and \textit{iv}) the $SHARPE$ value. It should be considered that for the computation of outcomes \textit{ii}, \textit{iii} and \textit{iv}, the Markowitz formulation of the $POP$ has been used as a base.

\section{\texttt{Q4FuturePOP}: Quantum Computing based System for Portfolio Optimization with Future Asset values and automatic universe reduction} \label{sec:Q4FuturePOP}

\begin{figure}[h!]
	\centering
	\begin{tabular}{cc}
		\includegraphics[width=1.0\columnwidth]{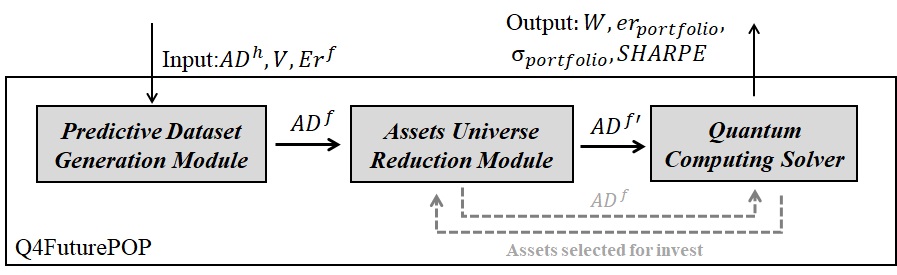}
	\end{tabular}
	\caption{A schematic description of the proposed \texttt{Q4FuturePOP}.}
	\label{fig:Q4FuturePOP}
\end{figure}

\texttt{Q4FuturePOP} consists of three interconnected modules: \texttt{PDG}, Assets Universe Reduction Module (\texttt{AUR}) and the Quantum Computing Solver Module (\texttt{QCS}). A schematic description of \texttt{Q4FuturePOP} is depicted in Figure \ref{fig:Q4FuturePOP}, in which the relationship of the three modules is represented.

In a nutshell, \texttt{PDG} is devoted to generating the complete dataset of future predicted values of $\mathcal{A}$. The second module, \texttt{AUR}, is in charge of intelligently reducing the complete $\mathcal{A}$ into $\mathcal{A'}$, while \texttt{QCS} is the module that solves the $POP$ and provides $W$ both for \texttt{AUR} or for the final user (as seen in Figure \ref{fig:Q4FuturePOP}). In the following subsections, we describe \texttt{PDG}, \texttt{AUR} and \texttt{QCS} in detail.

\subsection{Predicted Dataset Generation Module - \texttt{PDG}}\label{sec:generation}

To simulate future prices and eventually build $AD^{f}$, the \texttt{PDG} starts by generating a matrix $X\in\mathbb{R}^{{L-1}\times{N}}$ composed of random values drawn from a standard normal distribution, finding a matrix $A\in\mathbb{R}^{{L-1}\times{L-1}}$ and a bias $b\in\mathbb{R}^{{L-1}\times{N}}$ such that $er^{f} = AX + b$. For finding both $A$ and $b$, \texttt{PDG} follows two different procedures:

\begin{itemize}
	
	\item \textit{Finding $A$}. The Cholesky decomposition of the covariance matrix $Cov(X)$ and $Cov(er^{f})$ stands that there exist, respectively, two unique lower triangular matrix $L_{x}$ and $L_{h}$ such that $Cov(X)=L_{x}L_{x}^{T}$ and $Cov(er^{f})=L_{h}L_{h}^{T}$. Considering that $X$ is an invertible matrix, which is highly probable since the random vectors that make up $X$ are independent by construction, we set $A^T=L_{h}L^{-1}_{x}$. Thanks to this procedure, we meet our first constraint, as $Cov(er^{f})=Cov(er^{h})$.
	\item \textit{Finding $b$}. Let us consider $Y=AX$ and express the expected return as a function of the daily return. For an asset $a_{i}$ we have $ln(1 + Er^{f}_{i}) = \sum\limits_{k=1}^{L-1} ln(1 + y_{k, i} + b_{i})$ if and only if $\forall{k, i} \subseteq \mathbb{R}^{{L-1}\times{N}} y_{k, i} + b_{i} > -1 $. Then, we use the Taylor-Young expansion of $ln$ to find an approximation of $ln(1 + Er^{f}_{i})$ as a polynomial $P^{n}_{i}(x)$. If $\forall{k} \subseteq \mathbb{R}^{L-1}, |y_{k, i} + x| < 1 $ then $\lim\limits_{\substack{n \rightarrow +\infty}} P^{n}_{i}(x) = ln(1 + Er^{f}_{i})$. Now, we take $b_{i}$ as a real root of the polynomial $P^{n}_{i}(x) - ln(1 + Er^{f}_{i})$ that respects the preceding constraint. If this root does not exist, the \texttt{PDG} cannot find a good solution. So, the \texttt{PDG} repeats this for all the assets in order to obtain $b$.
	
\end{itemize}

After these values are calculated following this procedure, $AD^f$ is reconstructed from the initial values $V$ to the predicted $er^f$. 

\subsection{Quantum Computing Solver Module - \texttt{QCS}}\label{sec:QC}

Despite being the last module called along the workflow of \texttt{Q4FuturePOP}, it is appropriate to describe \texttt{QCS} here since it is employed also as part of the calculation made within \texttt{AUR}. In a nutshell, the \texttt{QCS} is the module in charge of taking a complete dataset of assets containing their daily values and solving the $POP$. We represent in Figure \ref{fig:QCS} a schematic description of \texttt{QCS}. As can be observed in that figure, this module is composed of three different components:

\begin{figure}[h!]
	\centering
	\begin{tabular}{cc}
		\includegraphics[width=1.0\columnwidth]{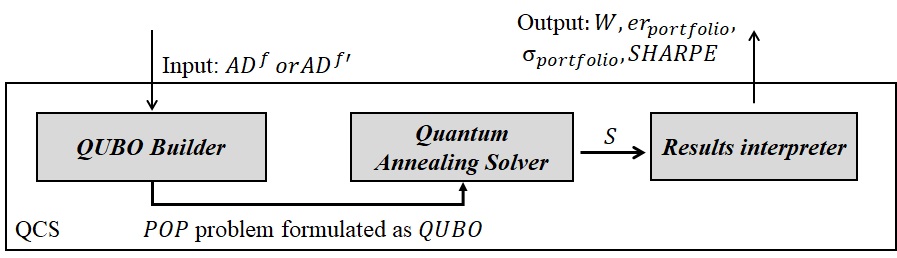}
	\end{tabular}
	\caption{A schematic description of \texttt{QCS}.}
	\label{fig:QCS}
\end{figure}

\textbf{\textit{QUBO Builder}}: the $POP$ dealt by \texttt{Q4FuturePOP} is tackled by a QC device. More specifically, the system is built to solve the problem by means of a quantum annealer. This specific type of device natively solves QUBO problems. For this reason, the first component of \texttt{QCS} has the objective of gathering the input dataset and modeling the $POP$ problem correctly. More specifically, for building the corresponding QUBO, we define the following Hamiltonian based on the formulation described in \cite{grant2021benchmarking}:

\begin{equation}
	\mathbf{H} = \alpha\mathbf{H_A} + \beta\mathbf{H_B} + \gamma\mathbf{H_C},
\end{equation}
where
\begin{equation}
	\mathbf{H_A} = \sum_{i}^{N-1}w_ier_i
\end{equation}
\begin{equation}
	\mathbf{H_B} = -\sum_{i,j}^{N-1}Cov_{i,j}x_ix_j.
\end{equation}
\begin{equation}
	\mathbf{H_C} = -(\sum_{i}^{n}w_i x_i - bd)^2.
\end{equation}
considering that $er_i$ represents the expected return for an asset $a_i$, and $x_i$ is a binary variable that is 1 if the asset $a_i$ has a $w_i>0$. Furthermore, the values $\alpha$, $\beta$ and $\gamma$ are float multipliers employed to weight each term.

\textbf{\textit{Quantum annealing solver}}: this is the central component of \texttt{QCS}, and where the call to the quantum device is made. As depicted in Figure \ref{fig:QCS}, \texttt{QCS} receives as input the problem modeled as a QUBO. This problem is tackled by a quantum annealer device, such as the ones provided by D-Wave: \texttt{Advantadge\_System6.1} or \texttt{Advantadge2\_prototype1.1}. Quantum-inspired alternatives, such as the Fujitsu Digital Annealer, are also eligible for being part of \texttt{QCS} as QUBO solver. Also, hybrid approaches such as Leap's hybrid Binary Quadratic Model Solver of D-Wave %, or Quantagonia's Hybrid Quantum Platform\footnote{https://www.quantagonia.com/}, 
can be embraced in this module. This component returns a list $S$ of binary values, which represents the solution to the QUBO ingested by the QC device.

\textbf{\textit{Results Interpreter}}: the solution $S$ provided by the quantum device is not directly interpretable for the final user. For this reason, this last component oversees the obtaining of the string of binary values provided by the quantum annealer and calculates the variables that will be returned as outcomes.

Thus, the \texttt{QCS} module is called in two different phases in the complete workflow of \texttt{Q4FuturePOP}. On the one hand, the \texttt{QCS} is called by \texttt{AUR} module in the process of asset universe reduction. In this iterative procedure, the \texttt{QCS} is executed $E$ different times, using as input the complete dataset $AD^f$. Through these repetitive runs, \texttt{AUR} aims to detect the most interesting assets for conducting a search space reduction of the $POP$ (more details in Section \ref{sec:AUR}). On the other hand, the \texttt{QCS} is called in the last stage of the complete \texttt{Q4FuturePOP} execution, using as input the reduced $AD^{f'}$, and with the goal of obtaining the final $W$, $er_{portfolio}$, $\sigma_{portfolio}$ and $SHARPE$ (calculated as $er_{portfolio}$/$\sigma_{portfolio}$).

\subsection{Assets Universe Reduction Module - \texttt{AUR}}\label{sec:AUR}

The motivation behind the implementation of the \texttt{AUR} module is to face the limitations of current quantum annealers. Despite all the research and developments made in the field, current quantum computers suffer from limitations such as a finite number of qubits or noisy processes that impact the performance of quantum solvers \cite{ajagekar2020quantum}. For this reason, the actual stage of QC field is known as NISQ era \cite{Preskill2018}.

Under this rationale, the main objective of \texttt{AUR} module is to decrease the complexity of the problem at hand by building a reduced sub-instance of the $POP$ problem. Thus, with a smaller solution space, the system is able to reach higher-quality and more robust solutions. The \texttt{AUR} works as follows:

\begin{enumerate}
	\item The \texttt{AUR} receives as input the complete set of data $AD^f$ generated by the previously described \texttt{PDG} module. 
	\item \texttt{AUR} solves the $POP$ problem using the \texttt{QCS} module (detailed in Section \ref{sec:QC}) and $AD^f$ as input. Despite \texttt{QCS} provides more information, it just stores the identifiers of the assets $a_i$ to which any proportion of $bd$ has been allocated. Thus, \texttt{AUR} generates a list $\mathcal{A'}$ which is a subgroup of $\mathcal{A}$ ($\mathcal{A'} \subseteq \mathcal{A}$).
	\item Until the number of execution conducted by \texttt{AUR} is less than $E$, step 2 is repeated.
	\item Once the process is finished, \texttt{AUR} builds a reduction of $AD$ considering the information of all the assets in $\mathcal{A'}$, and discarding all the data related to assets that have not been deemed in any of the $E$ executions conducted by step 2.
\end{enumerate}

\begin{figure}[h!]
	\centering
	\begin{tabular}{cc}
		\includegraphics[width=0.95\columnwidth]{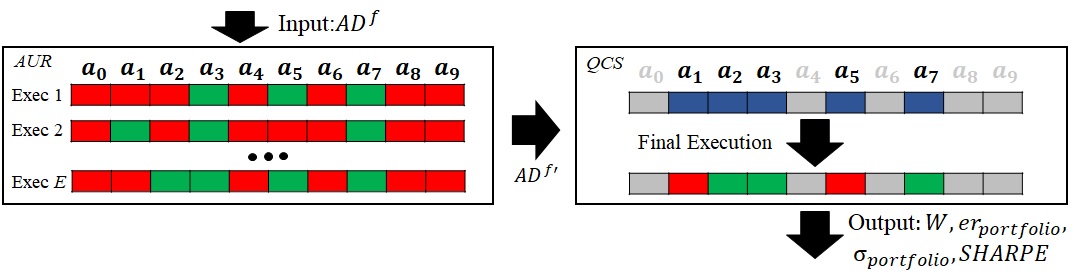}
	\end{tabular}
	\caption{A schematic description of \texttt{AUR} and its connection with \texttt{QCS}. \crule[Red]{0.3cm}{0.2cm} = asset with no budget allocated; \crule[ForestGreen]{0.3cm}{0.2cm} = asset with a budget $w_i>0$ allocated; \crule[Blue]{0.3cm}{0.2cm} = asset eligible for allocation in the reduced universe; \crule[Gray]{0.3cm}{0.2cm} = asset discarded in the universe reduction process}
	\label{fig:AUR}
\end{figure}

Lastly, for helping the understanding of \texttt{AUR} module, we depict a schematic description of \texttt{AUR} in Figure \ref{fig:AUR}, representing also the relation with \texttt{QCS} module.

\section{Discussion on the preliminary performance}\label{sec:discussion}

At this moment, \texttt{Q4FuturePOP} is in a prototypical stage of development, waiting to be completely validated as a whole system. Anyway, each module has been checked separately. On the one hand, the \texttt{PDG} has been successfully tested using a pool of predicted values provided by Welzia Management. In any case, due to extension constraints and the fact that this paper is more focused on QC, we will deepen the validation of the \texttt{PDG} module in a future research paper.

On the other hand, \texttt{AUR} and \texttt{QCS} modules have been jointly checked using a dataset also provided by Welzia Management. This dataset contains the daily values of 53 different assets over 12 years (from 01/01/2010 to 13/12/2022). Welzia Management also provided us with a set of historical portfolios chosen by the company's experts. Thus, using the dataset as input and the historical portfolios as baseline, six different use cases have been built for validation. Each of these instances consists of an excerpt of the complete dataset, with a depth ranging from 12 to 28 months. Therefore, for each use case, \texttt{AUR}+\texttt{QCS} modules of \texttt{Q4FuturePOP} have been run 6 independent times, and the results provided have been compared with the portfolios built by the experts. Also, it should be noted here that the quantum solver used for the conducted tests is the \texttt{Advantadge\_System6.2} of D-Wave, comprised of 5610 qubits and 40134 couplers spread over a Pegasus topology. 

The results of this preliminary experimentation are depicted in Table \ref{tab:results}, in which we represent the time frame that compose each dataset, the number of available assets, and the outcomes provided by both the experts and \texttt{AUR}+\texttt{QCS}. Each instance is coined as \texttt{UCX\_Y\_Z}, where $X$ is the ID of the dataset, $Y$ the time frame measured in months, and $Z$ the amount of assets deemed. Finally, it should be noted that both the employed datasets as well as the complete set of results obtained by \texttt{AUR}+\texttt{QCS} are available upon reasonable request.

Analyzing the results obtained by \texttt{AUR}+\texttt{QCS} modules, it should be noted that they have proved to be promising. These results have been approved by the experts from Welzia Management after several technical meetings. These meetings, and the fact of having the help of these experts, have been a really enriching point in the development of \texttt{Q4FuturePOP}. This is so since quantum-based solutions are usually analyzed from a purely academic point of view. That is, the solutions provided by the systems are usually analyzed based solely on their $SHARPE$ ratio. Although this ratio is a good indication of the quality of a solution, it does not represent the reality of the industry. High $SHARPE$ ratios can lead to triumphalist conclusions, which eventually clash with the reality of an industry with volatility that is difficult to perceive by a computer (whether quantum or classical). This is why an expert's judgment when generating a portfolio is an absolutely necessary factor.

\begin{table}[t!]
	\centering
	\caption{Results obtained from the preliminary experimentation carried out. Outcomes depicted for \texttt{AUR + QCS} are calculated using the averages obtained in the six independent runs. $er$ = expected return. $\sigma$ = risk.}
	\resizebox{0.9\columnwidth}{!}{
		\begin{tabular}{C{0.5in} C{1.0in} C{0.5in} | C{0.75in} C{0.5in} C{0.75in} C{0.75in}}
			\toprule[1.5pt]
			\multirow{2}{*}{\bf Instance} & \multirow{2}{*}{\bf Time Frame} & \multirow{2}{*}{\bf Assets} & \multicolumn{2}{c}{\bf Expert Results} & \multicolumn{2}{c}{\bf \texttt{AUR}+\texttt{QCS} Results}\\
			
			& & & $er$ & $\sigma$ & $er$ & $\sigma$\\
			
			\midrule
			\texttt{UC1\_12\_45}  & 31/01/2018 - 31/01/2019  & 45 & 22.56\% & \textbf{2.95} & \textbf{25.89\%} & 5.85\\
			\texttt{UC2\_12\_43}  & 05/05/2017 - 05/05/2018  & 43 & 12.95\% & \textbf{3.76} & \textbf{19,30\%} & 5.51\\
			\texttt{UC3\_24\_38}  & 01/01/2018 - 01/01/2020  & 35 & 14.05\% & \textbf{4.08} & \textbf{17,63\%} & 6.01\\
			\texttt{UC4\_28\_38}  & 24/01/2017 - 24/05/2019  & 38 & \textbf{11.64\%} & 8.99 & 8.62\% & \textbf{5.96}\\
			\texttt{UC5\_15\_40}  & 01/05/2018 - 01/08/2019  & 40 & \textbf{15.06}\% & 6.53 & 14.50\% & \textbf{4.12}\\		
			\texttt{UC6\_20\_53}  & 09/04/2021 - 09/12/2022  & 53 & 2.81\% & 12,31 & \textbf{6.37\%} & \textbf{11.62}\\
			
			\bottomrule[1.5pt]
		\end{tabular}
	}
	\label{tab:results}
\end{table}

All in all, the solutions achieved by the proposed system have proven to be promising, offering better results than the experts in some cases. In any case, as has been described, the fact that they present $SHARPE$ ratios higher than the portfolios proposed by Welzia Management does not imply that in practice they are better than the latter. Even so, Welzia Management has valued very positively the results obtained by \texttt{AUR}+\texttt{QCS}, being aware of the value that a platform of these characteristics can have in its day to day operation, acting as an assistant in their decision making processes.

\section{Conclusions and Future Work}\label{sec:conc}

In this paper, a quantum-based approach for solving the well-known Portfolio Optimization Problem has been presented, coined as \texttt{Q4FuturePOP}. Two are the main innovations inherent to the system proposed: \textit{i}) \texttt{Q4FuturePOP} is modeled for working with future prediction of assets instead of historical values; and \textit{ii}) \texttt{Q4FuturePOP} includes an \textit{automatic universe reduction} module, which is conceived to intelligently reduce the complexity of the problem. Along with the description of the system, we have also introduced a brief discussion about the preliminary performance of the different modules that compose the prototypical version of the tool.

Several research lines stem directly from the findings reported in this work. The first, and most obvious, is the validation of the complete system. Other future work includes the fine-tuning of the parameters that involve the generation of the POP's QUBO. Also, other quantum-based solvers apart from the ones provided by D-Wave are planned to be tested.

\section*{Acknowledgments}

This work was supported by the Spanish CDTI through Proyectos I+D Cervera 2021 Program (QOptimiza project, 095359). This work was also supported by the Basque Government through ELKARTEK program (BRTA-QUANTUM project, KK-2022/00041). The authors thank Miguel Uceda, Welzia’s Invesment Director, for his assistance and for providing the data employed for the tests conducted. 

\bibliographystyle{IEEEtran}
\bibliography{IEEEexample}

\end{document}